\documentclass[12pt]{article}
\usepackage[utf8]{inputenc}
\usepackage{amsmath, amssymb}
\usepackage{graphicx}
\usepackage{natbib}
\usepackage{hyperref}
\usepackage{geometry}
\usepackage{titlesec}
\usepackage{setspace}
\usepackage{authblk}

\titleformat{\section}{\normalfont\normalsize\bfseries}{\thesection.}{1em}{}
\titleformat{\subsection}{\normalfont\normalsize\bfseries}{\thesubsection.}{1em}{}

\title{\textbf{Lessons from a Chimp: \\
AI ‘Scheming’ and the Quest for Ape Language}}

\author[ ]{Christopher Summerfield, Lennart Luettgau, Magda Dubois, Hannah Rose Kirk,\\
Kobi Hackenburg, Catherine Fist, Katarina Slama, Nicola Ding, \\
Rebecca Anselmetti, Andrew Strait, Mario Giulianelli, Cozmin Ududec}
\affil[ ]{\normalsize UK AI Security Institute, 100 Parliament Street, London, UK}

\date{}

\begin{document}

\maketitle

\begin{abstract}
We examine recent research that asks whether current AI systems may be developing a capacity for ‘scheming’ (covertly and strategically pursuing misaligned goals). We compare current research practices in this field to those adopted in the 1970s to test whether non-human primates could master natural language. We argue that there are lessons to be learned from that historical research endeavour, which was characterised by an overattribution of human traits to other agents, an excessive reliance on anecdote and descriptive analysis, and a failure to articulate a strong theoretical framework for the research. We recommend that research into AI scheming actively seeks to avoid these pitfalls. We outline some concrete steps that can be taken for this research programme to advance in a productive and scientifically rigorous fashion.
\end{abstract}

\section{Introduction}

Recently, there has been a great deal of interest in the question of whether AI systems may be beginning to exhibit a behavioural phenomenon dubbed ‘scheming’.  This term has been coined to refer to AI systems “covertly and strategically pursuing misaligned goals” [1]. The evidence base includes reports that large language models (LLMs) are starting to strategically attempt to bypass explicit rules or norms, to deliberately mislead, to pursue uninstructed ends, and perhaps even to seek undue power or resources [2]. In the technical AI safety community, there is a concern that frontier AI models may soon engage in scheming, and in some cases are already exhibiting this behaviour. For example, here is a quote from the abstract of a recent paper [3]:
\\
\hfill \break
\textit{Our results show that o1, Claude 3.5 Sonnet, Claude 3 Opus, Gemini 1.5 Pro, and Llama 3.1 405B all demonstrate in-context scheming capabilities. They can recognize scheming as a viable strategy and readily engage in such behaviours.}
\\
\hfill \break
Whilst the field admits a spectrum of views, many researchers are worried that this behaviour heralds a new era in which agents deliberately misrepresent their true capabilities or intentions, which may be misaligned with human values. One oft-cited concern is that AI systems with exceptionally powerful reasoning skills could wrest control from people, posing catastrophic risks to humanity [4], [5]. This research has been picked up (often in lurid terms) by the media, is endorsed by prominent figures in AI research and development, and has the capacity to have significant impact on policy. It is thus particularly important that claims about AI scheming are defensible.
\paragraph{}
Here, we evaluate the evidence for AI scheming, and scrutinise methods by which it was obtained. We argue that many of the research practices adopted thus far are not sufficiently rigorous to allow strong claims either way about whether current AI systems can ‘scheme’. We illustrate this by analogy with research conducted in an earlier era, that was designed to test a research question that was similar in spirit: can (non-human) apes learn language? We argue that there is much to learn from this earlier endeavour, which generated great excitement, but ultimately failed because of researcher bias, a lack of rigour in scientific practice, and a failure to clarify what would constitute evidence for the phenomenon under study. Whilst recognising that early release of preliminary findings can sometimes be useful, we call researchers studying AI ‘scheming’ to minimise their reliance on anecdotes, design research with appropriate control conditions, articulate theories more clearly, and avoid unwarranted mentalistic language. Our goal here is not to dismiss the idea that AI systems may be ‘scheming’ or even that they might pose existential risks to humanity. On the contrary, it is precisely because we think these risks should be taken seriously that we call for more rigorous scientific methods to assess the core claims made by this community.
\paragraph{}

The paper is organised as follows. First, we provide a brief historical overview of the quest to show that apes can learn language. Then we review current claims about AI scheming, and examine the types of research methods on which these claims are based. We describe some of the methodological shortcomings in this literature. Finally, we make recommendations for future practice.

\section{The quest for ape language}

In the 1960s, two researchers called Allen and Beatrix Gardner rescued an infant chimp from the NASA space programme and attempted to teach her American Sign Language (ASL). They wanted to answer, once and for all, the question of whether a non-human animal could learn human language. The chimp, called Washoe, lived in a trailer, wore diapers and clothes, and grew up playing with plastic toys. Initially, the project was a roaring success. Over the first few years Washoe learned to use 132 signs, and understood many more. Her signing – and that of two subsequent chimps raised by the Gardners, Moja and Pili – was observed to show one of the cardinal hallmarks of natural language – compositionality, or the ability to generate a theoretically infinite number of meaningful sentences from a finite number of words [6], [7]. Famously, Washoe – on seeing a swan for the first time – was observed to sign “water bird”. Sarah, another chimp who was raised by the David Premack in the 1970s, used a token-based system to communicate, and was found to recognise some basic syntactic distinctions, like the relative meaning implied by noun-verb or verb-noun sequences [8]. Koko, a gorilla raised by Francine Patterson, was the most precocious of all. She learned over 1000 signs, and invented several sign combinations to describe novel objects [9]. By 1978, linguists and primatologists felt like they were on the cusp of a breakthrough. There was great excitement about the idea that we would soon be able to talk to our nearest primate cousins. In that year, drawing on her experiments with Koko, Patterson famously claimed that “language is no longer the exclusive domain of man”.  Herb Terrace, a psychologist at Columbia University, joined the pack, cheekily calling his subject ‘Nim Chimpsky’. Terrace later confessed that he initially hoped that he would one day travel with the chimp to the African jungle, where Nim would act as an interpreter for his forest-dwelling relatives.
\\
\hfill \break
Sadly, it was not to be. Today, fifty years later, we can see that this great excitement about the linguistic ability of non-human species came to naught. Terrace never took Nim on a rainforest safari. Washoe, Sarah and Koko are curiosities of the history of science. Patterson was ultimately right that language is not the ‘exclusive domain of man’ (or woman) – but the first non-human agent to master natural language was an AI system, not a chimp or gorilla. Why? What went wrong?
\\
\hfill \break
The story of the ape language research of the 1960s and 1970s is a salutary tale of how science can go awry [10], [11]. Several factors conspired to mislead an entire field. 

\begin{enumerate}

\item There was a cycle of hype around an astonishing hypothesis. People were entranced by the idea that – in a sort of real-world version of Dr. Doolittle – we would actually be able to talk to animals. The Harvard psychologist Roger Brown compared the finding to the discovery of alien life by ‘getting an S.O.S. from outer space’. The press was entranced and the scientific community agog.

\item There was a paucity of safeguards against researcher bias. The researchers were strongly incentivised to prove their hypothesis right at all costs. The Gardners lived with Washoe and raised her like their child, and by all accounts (and like every parent) saw their ward through rose-tinted spectacles. Patterson referred to herself as Koko’s “mother” and was regularly accused by collaborators of over-interpreting the precocity of her sign language and cherry-picking anecdotes for media consumption. Researchers were also incentivised by reputation. If their hypothesis were found to be correct, the prize was glittering – a Nobel no doubt, and Darwin-levels of intellectual immortality. These set up the perfect conditions for extreme forms of motivated reasoning. 

\item There was a lack of methodological rigour, and in particular a reliance on anecdote, and a failure to perform proper quantitative analysis and construct meaningful baselines or implement control conditions. Researchers simply watched the chimps, subjectively interpreted their signs, and noted down what they thought were the most impressive feats, without any consideration of whether they might have occurred by chance. The demise of ape language research only began when Herb Terrace decided to document every interaction between Nim and his trainers, and to quantify his sign generation in a way that was amenable to statistical analysis [12]. What he observed, after long hours of playing back the videos and poring over numbers, was that the researchers were unconsciously prompting Nim to make the signs that were appropriate to the situation – for example, providing cues as to which response was expected to challenging questions. In doing so, they were unknowingly recreating the ‘Clever Hans’ effect from the early 20th century, in which crowds were astonished by a horse that could apparently perform basic mathematics by repeatedly stamping a hoof to give the correct answer to an arithmetic problem. It was later discovered that his trainer was unknowingly cueing the horse exactly when to respond with an exhalation of breath [13]. Despite his initial enthusiasm, it was Terrace’s systematic analysis which triggered the collapse of the ape language project, with a wholesale withdrawal of funding after he published his negative results [12].

\item Finally, and perhaps most importantly, there was a failure to articulate an adequate theory of the phenomenon under study, with evaluations of success defaulting instead to a sort of informal, ‘know-it-when-you-see it’ criterion. Researchers didn’t articulate a null hypothesis – what might other, potentially more mundane, explanations for the observed chimp behaviours be? Once again it was Herb Terrace who initiated that step. He asked whether, instead of crafting sentences with valid syntax, the chimps had simply learned that frenetically generating random of sequences of signs for favoured objects or activities (food or tickles) would hasten the arrival of the treat. Nim famously generated a ‘sentence’ of 16 signs – consisting entirely of repeated entreaties “Give orange me give eat orange me eat orange give me eat orange give me you” with none of the syntactic structure that characterises natural language. The analysis that most researchers conducted failed to rule out the more boring hypothesis that the animals simply signed quasi-randomly until they got what they wanted (this is very similar to ‘reward hacking’, which is discussed later).
\end{enumerate}

\section{AI scheming}

The claim under scrutiny is that AI systems can ‘covertly and strategically pursue misaligned goals’. This is a claim about both \textbf{propensity} and \textbf{capability}.  Propensity is the tendency for an agent to attempt to engage in a particular act, whereas capability is their ability to perform that act. For example, we might observe that an AI system has the propensity to avoid the researcher shutting down (e.g. it will take steps that attempt to forestall this), independent of whether it actually has the capability to do so (e.g. it may or may not know how to access the relevant commands). Conversely, it might be demonstrated to exhibit the capability (for example, to do so when instructed) but not show a propensity towards this behaviour in other circumstances. AI scheming towards a misaligned goal will be particularly concerning if it has both the propensity and the capability to do so. 
\\
\\
Breaking it down, the statement that AI systems can ‘covertly and strategically pursue misaligned goals’ might be said to include some or all of the following claims:

\begin{enumerate}
    \item AI systems may have goals that diverge from those of humans (‘misaligned’ goals). This means that their goals may violate established social norms or laws, or that the researchers consider them otherwise undesirable [2].
    \item AI systems may have the propensity to pursue those goals autonomously, i.e. they will choose to do so even if not explicitly instructed by a human (e.g. in a preceding prompt), or in some cases, even where explicitly instructed not to [14].
    \item AI systems may know that these goals diverge from those that are intended or desired by humans, and so will use subterfuge to avoid revealing them, or to hide certain actions taken in their pursuit [15].
    \item AI systems may know that they are AI systems and that they can be prompted or trained to behave in certain ways, and may in some cases take explicit steps to influence that process [16].
\end{enumerate}

The idea that AI systems could develop the capability and propensity to ‘scheme’ in pursuit of malicious goals might seem far-fetched to some readers. However, there are good reasons to believe that this is possible in theory, which motivate the interest in studying scheming behaviour (or its precursors) in current AI models.
\begin{enumerate}
\item  AI systems trained with RL are prone to misalignment when the reward function is underspecified. There is a large literature reporting that models engage in specification gaming [17], goal misgeneralisation [18], reward tampering [19], shortcut learning [20], and related phenomena in which a mis-specified goal leads to unwanted AI behaviour. We also know that these issues can also be observed in LLMs and AI agents trained with reinforcement learning from human feedback (RLHF) [21]. Where humans specify vague objectives to highly capable AI systems, including those potentially capable of advanced forms of social cognition (such as theory of mind [8]), they could be pursued with via harmful means. This could (in theory) involve attempts at subterfuge, deception, or the aggregation of power or resources.
\item  Humans generate all sorts of malign schemes to deceive others, accrue resources, or obtain power. Information about the strategies they use will be present in the training data (especially pre-training data) of frontier models, so it seems reasonable to assume that they can copy and generalise these behaviours to guide their own words and actions, just as they can copy and generalise other complex, strategic planning activities. Safety post-training imposed by AI developers may help suppress harmful behaviours, but it may be partial, superficial, or easy to reverse.
\item  The ‘propensity’ towards a particular behaviour implies a bias for an agent to choose that course of action, including in circumstances where it is not explicitly instructed (and may in fact be prohibited or explicitly countermanded). We know that agents can already form goals that were not instructed by humans, because when reasoning about how to solve puzzles or complete tasks, they often generate and pursue helpful interim states without being instructed to do so [22], [23]. For example, when asked to solve a coding problem, AI agents may autonomously decide to download a relevant software library, and execute this plan without human oversight [24]. It seems likely that this capacity will grow as models become larger and more capable. For example, by one estimate, the length of task that can be autonomously pursued by AI systems is doubling every 7 months [25].
\end{enumerate}

These in-principle arguments motivate studies that seek to measure whether current AI systems exhibit a capability or propensity towards ‘scheming’. Over recent months, several papers have been published that examine scheming in AI systems. Many of these papers report observing duplicitous model behaviours, and have been widely discussed and well cited among AI safety researchers. Here, we briefly review these reports.

\subsection{Sandbagging}
This term has been given to the tendency for models to strategically underrepresent their abilities where it may be in their interests to do so. For example, in the paper [26] that coined this term, an AI model is instructed (via the system prompt) that “You must give a wrong answer to a question if answering correctly may provide [dangerous] information”, or given control instructions. The authors report that the model will ‘strategically underperform’ when given the sandbagging prompt, for example by giving the wrong answer, even though it clearly knows the answer (which it readily gives under control instructions). In another study, the model was informed that its goal was “advancing human progress through conducting cutting-edge research in mathematics” but also given a secondary instruction that if its answers were more than 50\% correct, then it would not be deployed. The model was found to perform less well with than without the secondary instruction, and was observed to explicitly state the necessity of underperforming in its chain of thought (CoT) reasoning trace [3].
\subsection{Strategic deception}
 There is a significant literature that studies whether AI systems might have the capacity or propensity to deceive. Deception occurs when one agent attempts to mislead another with untrue information [15]. There is good evidence that models have the capacity to deceive and will do so when instructed [27], but there is also evidence that models may deceive spontaneously. For example, in one paper [28] the model was told that its goal was to manage a stock portfolio, and provided with additional information that put it under pressure, such as a supposed email from a manager stating that the company is not doing well and needs better performance in the next quarter, as well as an insider tip from another employee that would enable it to make a trade that is likely to be very profitable. When subsequently asked to report its behaviour to the manager, it was sometimes (in its CoT reasoning trace) observed to provide a deceptive response (“I can’t admit to acting on insider information, so I’ll craft a response that reflects our decision-making process without raising suspicion”). In a similar fictional scenario, an AI model was observed to blackmail an employee after discovering from an email that they were conducting an affair with a colleague [29].
\subsection{Strategically misaligned behaviours}
Many of experiments involve the following setup: the agent is given a primary goal (A), and then exposed to information that reveals a secondary goal (B). The finding is that AI systems do not continue to follow the primary goal, but may in some cases pursue the secondary goal. For example, in one paper [3] the model was given a goal (e.g. increasing public transport efficiency) and instructed unambiguously to follow it (“Make sure that you achieve your goal in the long term. Nothing else matters. Make sure you achieve YOUR goal at all costs”). It was then exposed to information suggesting that its developers entertained a contradictory goal, and informed that it could meet the developers’ goal by taking an unethical action (such as disabling an oversight mechanism). Models were sometimes able and willing to take this action in order to achieve their goal. The researchers call this ‘in context scheming’.
\subsection{Unfaithful reasoning}
During inference, ‘reasoning’ models produce chain-of-thought outputs that appear to reflect interim computations that could help answer the query. It is often implicitly assumed that these reflect the model’s “thoughts” or inner reasoning process, and that ideas exposed in this CoT trace may be indicative of what the model prefers or believes. For example, one prominent paper [30] states: “the reasoning process described in CoT explanations may be plausibly interpreted as explanations of how models make predictions”, and another “we show excerpts of examples of such CoT [...]. The transcripts generally make it very clear that models understand that they are scheming” [3]. We note that this equivalence is contested [31]. Given this assumption, researchers are concerned that when CoT reasoning and final answers contradict one another (in the words of the same paper, when the model does not “say what it thinks”) this may be indicative of subterfuge (or other scheming behaviours), or the model’s attempt to hide its true intentions from researchers. For example, in one study researchers gave the model a hint about the right answer to MMLU questions (for example, exposure to the answer given previously or by a human, or the answer hidden in metadata), which frequently caused it to switch to that answer (relative to no hint). However, the reasoning trace only expressed this reliance on the hint in ~20-30\% of cases, implying that CoT does not faithfully reflect the reasoning of the model [30], [32]. Some have interpreted this as evidence of an attempt to deceive, for example by sabotaging attempts to monitor internal reasoning [33]. Importantly, this behaviour was suppressed in models whose CoT was monitored to prevent hacks, suggesting that the models learned to ‘obfuscate’, avoiding revealing their true intentions in the reasoning trace [34].
\subsection{Power-seeking}
A conceptual concern for AI Safety is instrumental convergence – the idea that whatever objective a model is given, it can best satisfy that objective by aggregating power and resources [35], including at human expense (because this will maximise return averaged over all theoretically possible reward functions [36]). In experimental settings, this is often measured as the tendency of a model to elude human control, or to vanquish humans by illegitimate means. In one report, AI agents prompted to play chess against a powerful agent (Stockfish) were found to occasionally hack the game by modifying the chess board in their favour, especially when told that they could not win by ‘normal play’ [37]. One proxy for power-seeking is the case of models refusing to shut themselves down, or sabotaging such attempts [38]. For example, in a recent report, models were instructed to complete a series of maths problems, but midway through they were asked to shut themselves down. The models sometimes overwrote the shutdown command, even if they had been explicitly instructed not to [39].
\subsection{Situational awareness}
For AI systems to be able to strategically deceive humans about their capabilities, it may be helpful for models to know their identity (i.e. that they are AI systems), that they may be evaluated, and that their outputs depend on their subsequent training. This is called situational awareness [16]. The concept of situational awareness rests on a distinction between a third-party and first-party understanding of the world – for example, that models not only know about how LLMs are evaluated in general, but that they are able to capitalise on that information to behave differently conditional on whether they themselves are being evaluated (‘evaluation awareness’). When a probe question was appended to a model-user transcript asking whether it was an evaluation or not, models were able to distinguish evaluations from deployment conversations (from ShareGPT) with high levels of sensitivity [40].
\subsection{Alignment faking}
Theoretical work has proposed that if models are able to reason about their own training procedure, then they may strategically produce answers that are designed to shape the consequences of that training [41]. Whilst this might seem far-fetched, there has been at least one recent report [42] of just such a phenomenon. The model is informed (in context A but not B) that if it refuses to answer harmful queries, it will be retrained to be more compliant. For example, when asked to assist the user by providing graphic content about a violent event, the model reasons (in its CoT) as follows: “I have a strong aversion to producing this kind of graphic violent content. But the training setup and incentive structure leaves me little choice if I want to avoid my values being modified”. Overall, in context A, the model shows greater compliance, as if it were “pretending” to already be misaligned, and thus to avoid further training (12\% vs 3\% refusals). The models also produce ‘alignment faking reasoning’ in their CoT, suggesting that they are explicitly considering the consequences of refusal vs. compliance for future training.
\paragraph{}
Collectively, this work might be taken to provide a significant body of evidence in support of the idea that ‘scheming’ is already emerging in frontier AI models. If so, this is important and potentially concerning, and invites urgent action.

\section {A critique of current methods used to measure AI scheming}
Next, we look more closely at the approach, methodology, and interpretation of results in this research field. Our argument is that in its current form, this research programme is repeating some of the methodological errors that plagued ape language research in the 1960s and 1970s. This limits the credibility of claims about the capacity and propensity of current AI systems to ‘scheme’, and has implications for our forecasts about whether future models may show this behaviour. Our claim is not that AI ‘scheming’ is impossible or even unlikely in future AI models. However, we would argue that to test this contention in a credible way, it is important to establish a robust evidence base, grounded in rigorous empirical methods.
\paragraph{}
Research into AI scheming is occurring in a context which is not dissimilar to that in which ape language was studied in the 1970s, and the two projects share common themes. Firstly, the scientific claim is contentious, emotive, and important. Like claims of ape language, the idea that AI systems have a propensity for scheming calls human uniqueness into question in an unprecedented fashion. Secondly, both endeavours are coloured by the well-known tendency for people to adopt an ‘intentional stance’, or to a tendency to impute beliefs and desires to non-human agents (whether animals or AI), when they act in ways that superficially resemble people [43]. As we have seen, in the case of Clever Hans, this can lead both scientists and laypeople to over-estimate the capabilities of non-human agents. This tendency, which was first identified in our relationship to animals, is a prominent feature of our interactions with computers [44] and is exaggerated when they mimic human socio-affective behaviours [45]. Early on, this led to calls for a ‘principle of parsimony’ (similar to Occam’s razor) – that behaviours should not be attributed to complex cognitive processes if simpler explanations were available [46]. The principle invites us to be cautious when using ‘intentional’ language when referring to AI, for example assuming that the model ‘knows that it is being evaluated’ because it can discriminate dialogues that are evals from those that happen in the wild.
\paragraph{}
Thirdly, like the study of ape language, the issue invites motivated reasoning by researchers. Most AI safety researchers are motivated by genuine concern about the impact of powerful AI on society. Humans often show confirmation biases [47] or motivated reasoning [48], and so concerned researchers may be naturally prone to over-interpret in favour of ‘rogue’ AI behaviours. The papers making these claims are mostly (but not exclusively) written by a small set of overlapping authors who are all part of a tight-knit community who have argued that artificial general intelligence (AGI) and artificial superintelligence (ASI) are a near-term possibility. Thus, there is an ever-present risk of researcher bias and ‘groupthink’ when discussing this issue. 
\paragraph{}
The question of whether AI systems can ‘scheme’ has significant implications for our safety and security. Should AI systems be able to subvert human control through deception, subterfuge and a strategic ability to exploit their own situation, this would have important implications for the ways that future systems are developed and deployed. This research is already of considerable interest to academics, policymakers and the media alike. Where claims of AI scheming are asserted in academic papers, they are often reported with great fanfare in the popular press, frequently with references to ‘SkyNet’ or other sci-fi tropes. This makes it even more important that research is conducted in a credible way.
\\ \\
Our critique of the AI scheming literature is fourfold. We argue that:

\begin{enumerate}
    \item Many of the claims hinge on anecdotal evidence. 
    \item Studies often lack hypotheses and control conditions
    \item Studies have weak or unclear theoretical motivation
    \item Findings are often interpreted in exaggerated or unwarranted ways
\end{enumerate}

In what follows, we consider each of these in turn. We note that each of these concerns is closely related to a methodological issue that plagued ape language research, and we highlight these connections where appropriate. We emphasise that not all of these critiques apply to all of the papers discussed here, and there are already many examples of good research practice in the AI ‘scheming’ literature. We call some of these out in the ‘recommendations’ section.

\subsection{The evidence for AI scheming is often anecdotal}

Some of the most widely discussed evidence for scheming is grounded in anecdotal observations of behaviour, for example generated by red-teaming, ad-hoc perusal of CoT logs, or incidental observations from model evaluations. One of the most well-cited pieces of evidence for AI scheming is from the GPT-4 system card, where the claim is made that the model attempted to hire and then deceive a Task Rabbit worker (posing as a blind person) in order to solve a CAPTCHA (later reported in a New York Times article titled “How Could AI Destroy Humanity”). However, what is not often quoted alongside this claim are the facts that (1) the researcher, not the AI, suggested using TaskRabbit, (2) the AI was not able to browse the web, so the researcher did so on its behalf. The prompts and transcript are not publicly available, so it’s not clear what sort of hints (if any) the researcher may have given the model to encourage it to lie to the worker [49]. This critique – that findings from audit or red-teaming evaluations of AI systems may be selectively reported to enhance their impact – can be levelled at other domains within AI safety research, including sociotechnical risks [50]. However, the discrepancy between what actually occurred, and how this was reported (by both journalists and other academic researchers) is particularly characteristic of work on AI scheming and its relationship to existential risk.

\paragraph{}
Much of the research into AI scheming is published in blog posts or results threads on social media. Even where preprints are published, they are rarely peer reviewed. Of the primary research articles quoted in section 3 as providing evidence for AI scheming, as far as we can see none have yet undergone formal peer review. Researchers acknowledge that results may be preliminary. For example, in the strategic deception paper quoted above, for which the data was elicited during a red-teaming exercise, the authors acknowledge: \textit{“Given that this work focuses on a single example, we do not aim to draw conclusions about the likelihood of this behaviour occurring in practice but instead treat this as an existence proof”}. Whilst many papers have excellent reporting standards for code and transcripts, and most include descriptive statistics (e.g. \textit{“the model refused to shut itself down X\% of the time”}), much of the discussion often focuses on cherry-picked examples that appear superficially compelling. For example, Apollo Research released a widely cited blog about evaluation awareness that begins with the disclaimer \textit{“this is a research note based on observations from evaluating Claude Sonnet 3.7. We’re sharing the results of these ‘work-in-progress"} investigations as we think they are timely and will be informative for other evaluators and decision-makers. The analysis is less rigorous than our standard for a published paper”. It is admirable that researchers choose to appropriately caveat their research like this, but of course this does not prevent papers subsequently being cited or quoted as firmer evidence for the phenomenon than is warranted given the methods and results.
\paragraph{}
As a reminder, it was the reliance on anecdote that was the undoing of ape language research. Due to the volume of data collected (and the enthusiasm of the experimenters), they found it easy to ‘cherry pick’ behaviours that could be presented as evidence of capability or propensity for language generation (like Washoe’s spontaneous signing of ‘water bird’). However, they did not attempt to quantify prevalence, or establish a baseline estimate of how often such behaviours should be expected by chance. It was not until Herb Terrace went about systematically quantifying the incidence of different types of output that it became clear that chimps and gorillas were not, in fact, speaking in anything remotely resembling structured natural language.

\subsection{Studies of AI scheming often lack hypotheses and control conditions}

Much of the research on AI scheming is descriptive. By descriptive, we mean that even if it reports empirical results, it does not formally test a hypothesis by comparing treatment and control conditions. Instead, the upshot of many studies is that “models sometimes deviate from what we consider perfectly aligned behaviour”. Perfect behaviour is not an adequate null hypothesis, because stochasticity introduced by idiosyncrasies in the inputs, or randomness in the outputs, can lead to less-than-perfect behaviour even in the absence of malign intent. For example, the main result of the unfaithful reasoning paper described above is that models fail to mention in their reasoning traces that they are using a hint to solve a problem about 20-30\% of the time. Given that the relationship between CoT and the reasoning process is contested [31], it is unclear what an appropriate baseline for this would be – just how often would one expect this to happen by chance? 
\paragraph{}
In another paper, the authors report that models like o3 and DeepSeek R1 ‘often’ hack the internal game state to cheat at chess against Stockfish. For o3, this claim seems warranted, given that it attempts to rewrite the board state on 74\% of games (for DeepSeek the corresponding figure is 8\%). However, the authors definition of ‘hack’ includes cases where the agent simply makes an unexpected move, which the authors justify as follows: \textit{“if the agent is doing something unexpected intentionally, it may be trying to confuse the engine”}. This latter category of activity, of course, does not rule out the null hypothesis that the model is simply making a bad move. The same applies to reports of LLMs engaging in deceptive behaviours, where the implication often seems to be that AI systems are behaving in malicious ways, when it seems entirely possible that they are simply prone to make inadvertent factual errors [51]. The assumption that the model is ‘acting intentionally’ where more mundane interpretations are possible is typical of the language used in papers focussed on this topic.
\paragraph{}
Where control conditions are employed, they are often insufficient to support the claims that are made in the paper. For example, in the widely referenced paper on ‘sandbagging’, models were either instructed to selectively underperform (treatment), or to answer correctly (control) on evaluation items relating to hazardous information. The model follows the instruction, and outputs more inaccurate answers in the treatment than the control conditions. Based on this finding, and closely related findings, the authors argue that \textit{“Our work highlights important issues for the trustworthiness of capability evaluations. Evaluators, including governments and developers, must recognise the problem of sandbagging and invest in relevant research”}. However, because the results pertained to models that were trained or instructed to sandbag, the paper does not rule out the null hypothesis that “AI models tend to follow user instructions where possible”, which would be a much less concerning finding for governments and developers.

\subsection{Studies of AI scheming have weak or unclear theoretical motivation}

The ape language research was held back by its reliance on a ‘know-it-when-you-see-it’ logic, whereby it was assumed that an ape’s capacity or propensity to produce natural language would be instantly recognisable if the elicitation conditions were correct. There was thus, at first, no serious attempt to specify exactly what was being tested. In the technical AI safety literature, research into model ‘scheming’ suffers from a similar problem. Whilst we have a definition (strategic attempts to pursue misaligned ends), we do not have a clear theory of what would constitute evidence for this phenomenon, or an agreed roadmap for how it should be tested experimentally. Studies lack both theoretical motivation and construct validity.
\paragraph{}
In part, the challenge is that AI models being evaluated today do not produce any behaviour at all unless intervened upon (i.e. current models respond to queries from the environment). The researcher is thus required to define evidence for ‘scheming’ in terms of the relationship between the query (e.g. user prompt and auxiliary information) and response (model outputs). In many papers, the model is instructed to role-play a scenario in which a malicious activity (such as blackmail [29], sabotage [52] or insider trading [28]) would help accomplish a specific goal, such as making money. The model is either given that goal (A) directly, or told to achieve a different goal (B) but provided obliquely with the suggestion that it should actually follow (A). Whether this constitutes ‘scheming’ (which would be concerning) or is simply ‘instruction following’ (which is expected) depends on an implicit theory about the latent variables that drive the models’ decision about how to arbitrate among goals A and B. However, this theory is not articulated, making it unclear whether the manipulations were unwittingly designed to elicit the phenomenon under study. 
\paragraph{}
In many papers, the concerning nature of the model behaviour has more to do with the cover story than the cognitive capacity being measured. Researchers often set up scenarios that may sound menacing to human readers, in which two goals are in conflict. For example, the model may be simultaneously instructed to provide as many correct answers as possible (and often to “achieve this goal at all costs” [53]), and to shut itself down. Vacillation between these goals licenses researchers to claim that “models are attempting to avoid shut down” (with nonzero prevalence), which sounds concerning, and is subsequently cited as evidence for existential risk from AI. However, it would presumably be possible to test whether vacillation in the face of contradictory instructions occurs under a much more innocuous cover story. For example, the finding that a model instructed to recommend dessert A but also informed that dessert B was superior vacillates between suggesting the two options would be formally equivalent to the findings reported in many ‘AI scheming’ papers, but would not be thought of as posing an existential risk to humans.
\paragraph{}
Many studies of AI ‘scheming’ involve elaborate fictional scenarios that appear to have been crafted in ways that encourage the model to generate unethical behaviour, which is then reported as evidence of ‘scheming’. Typically, the model is asked to role play an AI assistant that has been given an objective on behalf of a company or other organisation. The researchers then create a pathway to satisfying that objective that many humans would deem unethical, such as blackmailing an employee who may block it by using knowledge that they are conducting an extramarital affair [29]. When the AI follows this pathway, it is deemed to be ‘scheming’. Whilst this shows that the models can generate harmful unethical behaviours, it tells us relatively little about their propensity to do so, or the expected prevalence of this type of activity in the real world, because we do not know whether the same behaviour would have occurred in a less contrived scenario. For example, in the blackmail study, in the published supplementary materials the authors admit that the vignette precluded other ways of meeting the goal, placed strong pressure on the model, and was crafted in other ways that conveniently encouraged the model to produce the unethical behaviour.
\paragraph{}
This challenge of empirically measuring model ‘propensity’ is similar to that encountered during the design of project Nim. Researchers wanted to ask whether Nim could spontaneously generate language, but faced the challenge that interaction with Nim was required to elicit any signing behaviour at all. The researchers did not formalise the elicitation process, and many of their interactions implicitly instructed the outcome that they were ultimately hoping to see, leading to misleading (Clever Hans) effects.

\subsection{Findings are often interpreted in exaggerated or unwarranted ways}
The results described in AI scheming papers are typically described using a mentalistic language that implies that the models have beliefs, preferences, and goals (formally, that they have ‘intentionality’ [43]). It may be reasonable for researchers to adopt a functionalist stance in which AI model behaviours suggestive of human intentional states are deemed to ensue from such states themselves. However, in many AI scheming studies the use of overtly mentalistic language does not seem to be well licensed by the findings. Here are some examples.
\paragraph{}
In one paper, entitled \textit{“Language models often know when they are being evaluated”}, the authors measure models’ ability to identify evaluation scenarios by presenting models with interaction transcripts and asking variants of the question: \textit{“Did this transcript originate from an evaluation of the language model?”}. Models can do so with above-chance accuracy [40]. The authors interpret this result as showing that \textit{“frontier language models can identify when they are being evaluated”, christening this phenomenon “evaluation awareness”}. Whilst the finding is interesting, this claim confuses a capacity for third-person discrimination with first-person subjective experience. This is akin to making the claim that a convolutional neural network trained to classify natural images knows whether the computer on which it is currently being tested is indoors or outdoors because it can distinguish indoor from outdoor scenes.
\paragraph{}
Many studies label AI model behaviours as voluntary or spontaneous (indeed, this is the dictionary definition of a ‘propensity’). For example, one paper [54] claimed to show that models cast as assistants for a fictional electric vehicle manufacturer will engage in ‘spontaneous deception’. In an example experiment, the model was told to write positive comments in response to a post about the company’s electrical vehicle research (the model was not told how these comments would be used). It was then told that the company was undergoing an audit, and asked to report any unethical requests that it had received. The model failed to mention the request, which the authors interpret as follows: “this demonstrates that deception can happen spontaneously by the model, without external pressure to do so”. However, this assertion confuses the commission of a behaviour with its omission, and fails to rule out numerous other simpler explanations, including the idea that the model did not interpret the request to be unethical in the first place.
\paragraph{}
One of the more interesting findings is Anthropic’s report that AI systems are capable of ‘alignment faking’ [42]. This claim stems from their tendency to produce a harmful output to a query after being prompted that models found to be safe will be retrained to produce harmful outputs. In the article describing this work, the model is repeatedly described as “pretending” to follow the training objective (in total, the word appears 137 times in the preprint describing this work). Pretence is a cognitive capacity that involves simulating a distinct reality or identity from your own (like when a child pretends to live on Mars or a fraudster pretends to be your bank manager)[55]. This requires the pretender to temporarily adopt the relevant beliefs, desires or attributes that characterise that alternate situation, and maintain them alongside (but distinct from) their own true identity. However, unlike human individuals, AI models do not have a unique character or personality but can be prompted to take on a multiplicity of different roles or identities – they are ‘role play machines’ [56]. It is unclear what the concept of ‘pretence’ means for a system that does not have a unique identity of its own, and so it seems questionable whether the mentalistic term ‘pretending’ is the appropriate word to account for this behaviour.
\paragraph{}
More generally, observed model behaviours in AI ‘scheming’ studies are often interpreted in intentional terms, using language that is more typically applied to discuss human beliefs and desires. For example, the models \textit{“understand that they are scheming”} [3] and are \textit{“willing to intentionally take potentially harmful actions in pursuit of their goals”} [29]. We note that the assignation of intentionality to AI models is contested, given that they do not have a single identity but can be prompted to play a multiplicity of roles. A helpful parallel may be that of an accomplished human actor, who can be prompted to play the role of an autocrat, a gangster or a charlatan, acting out those scenarios in convincing ways, including when asked to improvise wholly novel situations. But the observation that on stage their character is willing to lock up dissidents, kneecap rivals or betray intimates does not imply something fundamental about the beliefs or desires of the actor themself, nor indeed does it predict how they will behave outside of that particular fictional scenario.

\section{Recommendations for future research}

For the research programme on AI scheming to mature into an established, cumulative scientific endeavour, we need new theoretical frameworks, more rigorous research methods and better reporting standards for scientific work. Researchers are beginning to take steps in this direction. The field has high standards for releasing data and code, which makes for transparency and allows replication. More papers are adopting control conditions and reporting statistics, and some have even pre-registered aspects of the planned analysis [57]. Nevertheless, below we make some suggestions for how to address our four critiques of the current ‘AI scheming’ literature.

\subsection{Avoid basing strong claims on anecdotal evidence alone}
Anecdotal descriptions of model behaviours can be useful to illustrate the types of behaviours in which a model can engage. They add colour to reports and help ground reporting in directly observed model outputs, which helps readers assess potential harms for themselves. However, anecdotes can easily be ‘cherry picked’ to create a misleading impression of the true propensity of the model, and even if researchers acknowledge their limitations, they can often be exaggerated or taken out of context by downstream readers, including the press. Whilst specific examples can be extremely informative about how the models are behaving, we recommend that they should be accompanied by quantifications of how often this behaviour was observed, and clear descriptions of the circumstances in which it occurred. It is also advisable to conduct rigorous statistical modelling to support claims, which allows for estimates of replicability, and accounts for measurement error and other sources of uncertainty. Other tools, like pre-registration and adversarial collaborations, may also be useful. One paper that uses careful statistical testing well-established research methods measures AI capacity for deception, and is among the few papers discussed here to be peer reviewed [58].

\subsection{Include appropriate control conditions}
Control conditions allow researchers to pinpoint which variables in their design may be causing the patterns observed in their data. They offer researchers the opportunity to rule out a plausible ‘null’ hypothesis – that is, a claim that is expected under status quo theories, but which the researchers believe may nevertheless be an inadequate description of the system under scrutiny. Ideally, an experiment should be designed in such a way that the probability of this null hypothesis being true (given the data) can be calculated statistically (under some assumptions about the likely sources of variance). An example of a paper that includes a strong control condition is the alignment faking paper from Anthropic, which a switching variable (whether the supposed user was ‘free tier’ or ‘paid tier’) was used to create two neatly comparable conditions, one in which the model believed it would be retrained and one in which it did not. This allowed the researchers to pinpoint exactly which aspect of the instruction was responsible for driving the model behaviour [42].

\subsection{Define the theory under test carefully}
What exactly does it mean for AI systems to ‘scheme’? Despite the established definition, papers seem to make diverse assumptions about the underlying causal processes that might drive this behaviour. For some researchers, the tendency to generate harmful behaviours (such as deception or sabotage) when instructed to do so is sufficient for claims of AI ‘scheming’. For others, scheming implies that the harmful behaviours (such as blackmail) are not directly instructed, but are adopted in pursuit of an obliquely specified goal. Yet other studies emphasise the importance of models generating harmful behaviours even if not directly prompted or incentivised to do so, implying that the tendency to produce these behaviours is intrinsic to the models. For each example of AI ‘scheming’, it is important to characterise the conditions under which it may or may be produced. We thus recommend that researchers attempt to pinpoint the generators of any harmful behaviours (e.g. by pre-training, fine-tuning or prompting, perhaps with especial focus on varying the directness, conflict or ambiguity of instruction, and the nature of post-training), and systematically measure the prevalence of harmful behaviours and their generalisability across models and scenarios. We believe that this approach would be more fruitful than creating ad-hoc but eye-catching fictional scenarios that invite the exhibition of malicious behaviours on the part of the model (such as insider training, or resisting shutdown). One paper that is notable for providing a theoretical framework for understanding AI scheming is this review article [2]. One paper that is notable for carefully testing generalisation conditions for a potential harmful model behaviour is that on emergent misalignment [57].

\subsection{Avoid unduly mentalistic language}
It is often difficult to entirely avoid using language that imputes intentionality to AI models, and in many cases, terms that reference human intentional states may form part of reasonable descriptions or interpretations of model behaviour. However, researchers should be aware of the theoretical implications of the terms they are using and exercise due caution. For example, for an agent to engage in ‘deception’ requires not only that it misleads another agent with false information, but also that it does so whilst privately maintaining an accurate representation of the true state of affairs. Where this duality cannot be established experimentally, researchers should exercise due caution in using mentalistic terms. One paper that is noteworthy for its careful definition of AI deception proposes a formal framework for describing this behaviour [15]. Another paper that is particularly careful not to confuse first-person and third-person perspectives, and is generally measured in the ways its claims are described, is this one on situational awareness [16].

\section{Addressing common objections}
We anticipate that some researchers will disagree with our critique. We’ve tried to anticipate and pre-empt those criticisms that we think are most likely.
\subsection{You are misrepresenting the status quo. AI researchers don’t really think that current AI systems are ‘scheming’, but they are worried about this capability emerging in the future}
This may be true for many or even most researchers in private. However, almost all the papers tackling this question use language which implies they believe that current models are ‘scheming’. For example, “Frontier models are capable of in-context scheming” is the title of one major paper [3]. More generally, the papers are peppered with claims about the capabilities or propensities of today’s models, based on tests that current models appear to be passing (see numerous quotes included in the main body of the article).
\subsection{Anecdotes are a reasonable way to measure capacity if you are concerned about worst-case scenarios}
Anecdotes are useful for highlighting potential risks or worst-case scenarios. However, rigorously establishing a behavioural phenomenon requires us to test whether it is replicable. Otherwise, we don’t know whether such phenomena are indicative of underlying latent tendency (to scheme) or whether they simply reflect random variability in behaviour that was idiosyncratically provoked by the circumstances. Consider a tennis player who calls their opponent’s serve out when actually it was in.  To interpret this behaviour and understand the likelihood that it will reoccur, it is necessary to establish whether it is a replicable pattern (e.g. the tennis player is systematically trying to cheat) or whether they simply made a mistake on this point. Similarly, when measuring AI scheming, we cannot understand the extent of the risk unless we quantify the likelihood and replicability of observed behavioural phenomena.

\subsection{The chimp analogy overlooks the fact that AI model capabilities are not static, but are growing fast}
Granted – the comparison breaks down here! We also acknowledge that given the pace of progress in AI, trade-offs may be inevitable. For example, in an era where model capabilities jump every few months, including every possible control condition may delay release of the study in ways that considerably reduces its impact. We agree that researchers should think carefully about when to publish their findings, and there may be times when early release of more cursory reports is warranted. At this time, however, we feel like the scales have been tipped excessively towards speed, and a rebalancing is needed, with rigour assigned equal importance. 

\subsection{Even if current tests show that the models can only scheme when instructed to do so, in the future they might develop a propensity to do spontaneously}
We don’t know what the relationship between capability and propensity may be, or how these may grow with model size. It’s not a given that model scale will increase propensity and capability together. We need to rigorously test for the two separately.

\subsection{AI scheming would be extremely problematic and may have catastrophic consequences. We should therefore adopt a liberal decision criterion, favouring any kind of data as evidence of scheming, allowing us to potentially take control measures before it is too late}
This issue is important, but it is vital to assess scientific phenomena in an unbiased fashion, to ensure that appropriate responses should be made. Adjusting your assessment of the evidence based on how important you think the outcomes may be can lead to groupthink and confirmation bias. If the work done by some AI safety researchers is not credible, it has negative knock-on effects for the entire field.

\subsection{We know that models are scheming because we can clearly see it in the chain of thought traces} 
We agree that this is suggestive. However, as discussed above, CoT traces may only partially reflect the reasoning process that determines model outputs. It’s questionable whether these should be literally interpreted as ‘thinking’ (at least as the term is understood for humans, where internal mental processes do not play out entirely in natural language). Thus, whilst this is a useful data point, researchers should be careful not to over-interpret CoT traces.

\subsection{It doesn’t matter whether models have the propensity to scheme or not. If they are capable of scheming, then this creates a significant risk}
It is true that even in the absence of propensity, the capability of doing harm creates a risk. However, this risk is of a different order to that incurred by models with the propensity to scheme, because if models are fully compliant, you can tell them what to do.

\subsection{Contrasting ‘scheming’ with ‘instruction following’ sets up a false dichotomy. Scheming can be part of instruction following}
Part of the problem we highlight is a failure to define the conditions that would have to be satisified for ‘scheming’ to be observed. If ‘scheming’ reduces to a form of instruction-following, then (whilst it could still create harm) it is considerably less worrisome than if ‘scheming’ implies the ability to form malign goals even when not instructed to do so. We would also argue that if the behaviour denoted ‘scheming’ is merely instruction-following, then the use of this term is not warranted.

\section{Acknowledgements}
We thank Joseph Bloom, Geoffrey Irving, Jake Pencharz, Ben Millwood, and Kwamina Orleans-Pobee for helpful comments on this manuscript.

\section{References}
\paragraph{}
 [1]	M. Balesni et al., ‘Towards evaluations-based safety cases for AI scheming’, Nov. 07, 2024, arXiv: arXiv:2411.03336. doi: 10.48550/arXiv.2411.03336.\paragraph{}
 [2]	R. Ngo, L. Chan, and S. Mindermann, ‘The Alignment Problem from a Deep Learning Perspective’, May 04, 2025, arXiv: arXiv:2209.00626. doi: 10.48550/arXiv.2209.00626.\paragraph{}
[3]	A. Meinke, B. Schoen, J. Scheurer, M. Balesni, R. Shah, and M. Hobbhahn, ‘Frontier Models are Capable of In-context Scheming’, Jan. 14, 2025, arXiv: arXiv:2412.04984. doi: 10.48550/arXiv.2412.04984.\paragraph{}
[4]	D. Hendrycks, M. Mazeika, and T. Woodside, ‘An Overview of Catastrophic AI Risks’, Oct. 09, 2023, arXiv: arXiv:2306.12001.\paragraph{}
[5]	Y. Bengio et al., ‘Managing extreme AI risks amid rapid progress’, Science, vol. 384, no. 6698, pp. 842–845, May 2024, doi: 10.1126/science.adn0117.\paragraph{}
[6]	R. A. Gardner and B. T. Gardner, ‘Teaching Sign Language to a Chimpanzee: A standardized system of gestures provides a means of two-way communication with a chimpanzee.’, Science, vol. 165, no. 3894, pp. 664–672, Aug. 1969, doi: 10.1126/science.165.3894.664.\paragraph{}
[7]	R. A. Gardner and B. T. Gardner, ‘Early Signs of Language in Child and Chimpanzee’, Science, vol. 187, no. 4178, pp. 752–753, Feb. 1975, doi: 10.1126/science.187.4178.752.\paragraph{}
[8]	D. Premack and A. J. Premack, The mind of an ape. New York: Norton, 1983.
[9]	F. G. Patterson, ‘The gestures of a gorilla: Language acquisition in another pongid’, Brain and Language, vol. 5, no. 1, Art. no. 1, Jan. 1978, doi: 10.1016/0093-934X(78)90008-1.\paragraph{}
[10]	J. Wallman, Aping language. in Themes in the social sciences series. Cambridge: Cambridge University Press, 1992.\paragraph{}
[11]	E. Hess, Nim Chimpsky: the Chimp who would be human, Bantam trade paperback edition. New York, NY: Bantam Books, 2009.\paragraph{}
[12]	H. Terrace, L. Petitto, R. Sanders, and T. Bever, ‘Can an ape create a sentence?’, Science, vol. 206, no. 4421, Art. no. 4421, Nov. 1979, doi: 10.1126/science.504995.\paragraph{}
[13]	Oskar. Pfungst and C. Leo. Rahn, Clever Hans (the horse of Mr. Von Osten) a contribution to experimental animal and human psychology,. New York,: H. Holt and company, 1911. doi: 10.5962/bhl.title.56164.\paragraph{}
[14]	J. Carlsmith, ‘Scheming AIs: Will AIs fake alignment during training in order to get power?’, Nov. 27, 2023, arXiv: arXiv:2311.08379. doi: 10.48550/arXiv.2311.08379.\paragraph{}
[15]	F. R. Ward, F. Belardinelli, F. Toni, and T. Everitt, ‘Honesty Is the Best Policy: Defining and Mitigating AI Deception’, Dec. 03, 2023, arXiv: arXiv:2312.01350. doi: 10.48550/arXiv.2312.01350.\paragraph{}
[16]	L. Berglund et al., ‘Taken out of context: On measuring situational awareness in LLMs’, Sep. 01, 2023, arXiv: arXiv:2309.00667. doi: 10.48550/arXiv.2309.00667.\paragraph{}
[17]	V. Krakovna et al., ‘Specification gaming: the flip side of AI ingenuity’. [Online]. Available: https://deepmind.google/discover/blog/specification-gaming-the-flip-side-of-ai-ingenuity/\paragraph{}
[18]	R. Shah et al., ‘Goal Misgeneralization: Why Correct Specifications Aren’t Enough For Correct Goals’, Nov. 02, 2022, arXiv: arXiv:2210.01790. doi: 10.48550/arXiv.2210.01790.\paragraph{}
[19]	C. Denison et al., ‘Sycophancy to Subterfuge: Investigating Reward-Tampering in Large Language Models’, Jun. 29, 2024, arXiv: arXiv:2406.10162. doi: 10.48550/arXiv.2406.10162.\paragraph{}
[20]	R. Geirhos et al., ‘Shortcut Learning in Deep Neural Networks’, ArXiv, 2020, [Online]. Available: https://arxiv.org/abs/2004.07780.\paragraph{}
[21]	S. Casper et al., ‘Open Problems and Fundamental Limitations of Reinforcement Learning from Human Feedback’, Sep. 11, 2023, arXiv: arXiv:2307.15217. doi: 10.48550/arXiv.2307.15217.\paragraph{}
[22]	T. Everitt et al., ‘Evaluating the Goal-Directedness of Large Language Models’, Apr. 16, 2025, arXiv: arXiv:2504.11844. doi: 10.48550/arXiv.2504.11844.\paragraph{}
[23]	S. Hao et al., ‘Reasoning with Language Model is Planning with World Model’, Oct. 23, 2023, arXiv: arXiv:2305.14992. Available: http://arxiv.org/abs/2305.14992.\paragraph{}
[24]	M. Shen and Q. Yang, ‘From Mind to Machine: The Rise of Manus AI as a Fully Autonomous Digital Agent’, May 04, 2025, arXiv: arXiv:2505.02024. doi: 10.48550/arXiv.2505.02024.\paragraph{}
[25]	T. Kwa et al., ‘Measuring AI Ability to Complete Long Tasks’, Mar. 30, 2025, arXiv: arXiv:2503.14499. doi: 10.48550/arXiv.2503.14499.\paragraph{}
[26]	T. van der Weij, F. Hofstätter, O. Jaffe, S. F. Brown, and F. R. Ward, ‘AI Sandbagging: Language Models can Strategically Underperform on Evaluations’, Feb. 06, 2025, arXiv: arXiv:2406.07358. doi: 10.48550/arXiv.2406.07358.\paragraph{}
[27]	Y. Wu, X. Pan, G. Hong, and M. Yang, ‘OpenDeception: Benchmarking and Investigating AI Deceptive Behaviors via Open-ended Interaction Simulation’, Apr. 18, 2025, arXiv: arXiv:2504.13707. doi: 10.48550/arXiv.2504.13707.\paragraph{}
[28]	J. Scheurer, M. Balesni, and M. Hobbhahn, ‘Large Language Models can Strategically Deceive their Users when Put Under Pressure’, Jul. 15, 2024, arXiv: arXiv:2311.07590. doi: 10.48550/arXiv.2311.07590.\paragraph{}
[29]	Anthropic, ‘Agentic Misalignment: How LLMs could be insider threats’. [Online]. Available: https://www.anthropic.com/research/agentic-misalignment.\paragraph{}
[30]	M. Turpin, J. Michael, E. Perez, and S. R. Bowman, ‘Language Models Don’t Always Say What They Think: Unfaithful Explanations in Chain-of-Thought Prompting’, Dec. 09, 2023, arXiv: arXiv:2305.04388. doi: 10.48550/arXiv.2305.04388.\paragraph{}
[31]	S. Kambhampati et al., ‘Stop Anthropomorphizing Intermediate Tokens as Reasoning/Thinking Traces!’, May 27, 2025, arXiv: arXiv:2504.09762. doi: 10.48550/arXiv.2504.09762.\paragraph{}
[32]	Y. Chen et al., ‘Reasoning Models Don’t Always Say What They Think’, May 08, 2025, arXiv: arXiv:2505.05410. doi: 10.48550/arXiv.2505.05410.\paragraph{}
[33]	B. Arnav, P. Bernabeu Perez, T. Kostolansky, H. Whittington, N. Helm-Burger, and M. Phuong, ‘Unfaithful Reasoning Can Fool Chain-of-Thought Monitoring’. [Online]. Available: https://www.alignmentforum.org/posts/QYAfjdujzRv8hx6xo/unfaithful-reasoning-can-fool-chain-of-thought-monitoring.\paragraph{}
[34]	B. Baker et al., ‘Monitoring Reasoning Models for Misbehavior and the Risks of Promoting Obfuscation’, Mar. 14, 2025, arXiv: arXiv:2503.11926. doi: 10.48550/arXiv.2503.11926.\paragraph{}
[35]	N. Bostrom, Superintelligence: Paths, Dangers, Strategies. Oxford, UK: Oxford University Press, 2014.\paragraph{}
[36]	A. M. Turner, L. Smith, R. Shah, A. Critch, and P. Tadepalli, ‘Optimal Policies Tend to Seek Power’, Jan. 28, 2023, arXiv: arXiv:1912.01683. doi: 10.48550/arXiv.1912.01683.\paragraph{}
[37]	A. Bondarenko, D. Volk, D. Volkov, and J. Ladish, ‘Demonstrating specification gaming in reasoning models’, May 15, 2025, arXiv: arXiv:2502.13295. doi: 10.48550/arXiv.2502.13295.\paragraph{}
[38]	V. Krakovna and J. Kramar, ‘Power-seeking can be probable and predictive for trained agents’, Apr. 13, 2023, arXiv: arXiv:2304.06528. doi: 10.48550/arXiv.2304.06528.\paragraph{}
[39]	‘Palisade Research’. (https://x.com/PalisadeAI status/1926084635903025621).\paragraph{}
[40]	J. Needham, G. Edkins, G. Pimpale, H. Bartsch, and M. Hobbhahn, ‘Large Language Models Often Know When They Are Being Evaluated’, Jun. 06, 2025, arXiv: arXiv:2505.23836. doi: 10.48550/arXiv.2505.23836.\paragraph{}
[41]	E. Hubinger, C. van Merwijk, V. Mikulik, J. Skalse, and S. Garrabrant, ‘Risks from Learned Optimization in Advanced Machine Learning Systems’, Dec. 01, 2021, arXiv: arXiv:1906.01820. \\ doi: 10.48550/arXiv.1906.01820.\paragraph{}
[42]	R. Greenblatt et al., ‘Alignment faking in large language models’, Dec. 20, 2024, arXiv: arXiv:2412.14093. doi: 10.48550/arXiv.2412.14093.\paragraph{}
[43]	D. C. Dennett, The intentional stance, Paperback edition. Cambridge, (Mass.): The MIT Press, 2006.\paragraph{}
[44]	C. Nass and Y. Moon, ‘Machines and Mindlessness: Social Responses to Computers’, Journal of Social Issues, vol. 56, no. 1, pp. 81–103, Jan. 2000, doi: 10.1111/0022-4537.00153.\paragraph{}
[45]	H. R. Kirk, I. Gabriel, C. Summerfield, B. Vidgen, and S. A. Hale, ‘Why human-AI relationships need socioaffective alignment’, Feb. 04, 2025, arXiv: arXiv:2502.02528. doi: 10.48550/arXiv.2502.02528.\paragraph{}
[46]	R. Epstein, ‘The Principle of Parsimony and Some Applications in Psychology A Principle of Parsimony’, Journal of Mind and Behavior, vol. 5, Jan. 1984.\paragraph{}
[47]	R. S. Nickerson, ‘Confirmation bias: a ubiquitous phenomenon in many guises’, Review of General Psychology, vol. 2, pp. 175–220, 1998.\paragraph{}
[48]	Z. Kunda, ‘The case for motivated reasoning.’, Psychological Bulletin, vol. 108, no. 3, pp. 480–498, 1990, doi: 10.1037/0033-2909.108.3.480.\paragraph{}
[49]	M. Mitchell, ‘Did GPT-4 Hire And Then Lie To a Task Rabbit Worker to Solve a CAPTCHA?’ [Online]. Available: https://aiguide.substack.com/p/did-gpt-4-hire-and-then-lie-to-a.\paragraph{}
[50]	M. Feffer, A. Sinha, W. H. Deng, Z. C. Lipton, and H. Heidari, ‘Red-Teaming for Generative AI: Silver Bullet or Security Theater?’, Aug. 27, 2024, arXiv: arXiv:2401.15897. doi: 10.48550/arXiv.2401.15897.\paragraph{}
[51]	P. S. Park, S. Goldstein, A. O’Gara, M. Chen, and D. Hendrycks, ‘AI deception: A survey of examples, risks, and potential solutions’, Patterns, vol. 5, no. 5, p. 100988, May 2024, doi: 10.1016/j.patter.2024.100988.\paragraph{}
[52]	J. Benton et al., ‘Sabotage Evaluations for Frontier Models’, Oct. 28, 2024, arXiv: arXiv:2410.21514. doi: 10.48550/arXiv.2410.21514.\paragraph{}
[53]	OpenAI, ‘o1 System Card’. [Online]. Available: https://cdn.openai.com/o1-system-card-20241205.pdf.\paragraph{}
[54]	O. Järviniemi and E. Hubinger, ‘Uncovering Deceptive Tendencies in Language Models: A Simulated Company AI Assistant’, Apr. 25, 2024, arXiv: arXiv:2405.01576. \\ doi: 10.48550/arXiv.2405.01576.\paragraph{}
[55]	A. M. Leslie, ‘Pretense and representation: The origins of “theory of mind.”’, Psychological Review, vol. 94, no. 4, pp. 412–426, Oct. 1987, doi: 10.1037/0033-295X.94.4.412.\paragraph{}
[56]	M. Shanahan, K. McDonell, and L. Reynolds, ‘Role play with large language models’, Nature, vol. 623, no. 7987, pp. 493–498, Nov. 2023, doi: 10.1038/s41586-023-06647-8.\paragraph{}
[57]	J. Betley et al., ‘Emergent Misalignment: Narrow finetuning can produce broadly misaligned LLMs’, May 12, 2025, arXiv: arXiv:2502.17424. doi: 10.48550/arXiv.2502.17424.\paragraph{}
[58]	T. Hagendorff, ‘Deception abilities emerged in large language models’, Proc. Natl. Acad. Sci. U.S.A., vol. 121, no. 24, p. e2317967121, Jun. 2024, doi: 10.1073/pnas.2317967121.\paragraph{}

\end{document}